\documentclass{article}
\usepackage[preprint]{spconf}
\usepackage{amssymb,amsmath,amsfonts}
\usepackage{booktabs}
\usepackage{multirow}
\usepackage{hyperref}
\usepackage{algorithm}
\usepackage{algpseudocode}
\usepackage{array}
\usepackage{graphicx}
\usepackage{tabularx}
\usepackage{nth}
\usepackage{xcolor}
\usepackage{dsfont}
\newcolumntype{Y}{>{\centering\arraybackslash}X}
\DeclareMathOperator*{\argmax}{argmax}

\title{Camera Model Identification with SPAIR-Swin and Entropy based Non-Homogeneous Patches}
\name{
\begin{tabular}{c}
Protyay Dey$^{1}$, Rejoy Chakraborty$^{1}$, Abhilasha S. Jadhav$^{1}$, Kapil Rana$^{2}$ \\
Gaurav Sharma$^{3}$, Puneet Goyal$^{1,4}$
\end{tabular}
}

\address{$^{1}$Computer Science and Engineering, Indian Institute of Technology Ropar, Punjab, India \\
$^{2}$Computer Science and Engineering, Thapar Institute of Engineering and Technology, Punjab, India \\
$^{3}$Electrical and Computer Engineering, University of Rochester, Rochester, New York, USA \\ 
$^{4}$Artificial Intelligence \& Data Engineering, Indian Institute of Technology Ropar, Punjab, India}

\begin{document}
%
\maketitle

\begin{abstract}
Source camera model identification (SCMI) plays a pivotal role in image forensics with applications including authenticity verification and copyright protection. For identifying the camera model used to capture a given image, we propose SPAIR-Swin, a novel model combining a modified spatial attention mechanism and inverted residual block (SPAIR) with a Swin Transformer. SPAIR-Swin effectively captures both global and local features, enabling robust identification of artifacts such as noise patterns that are particularly effective for SCMI. Additionally, unlike conventional methods focusing on homogeneous patches, we propose a patch selection strategy for SCMI that emphasizes high-entropy regions rich in patterns and textures. Extensive evaluations on four benchmark SCMI datasets demonstrate that SPAIR-Swin outperforms existing methods, achieving patch-level accuracies of 99.45\%, 98.39\%, 99.45\%, and 97.46\% and image-level accuracies of 99.87\%, 99.32\%, 100\%, and 98.61\% on the Dresden, Vision, Forchheim, and Socrates datasets, respectively. Our findings highlight that high-entropy patches, which contain high-frequency information such as edge sharpness, noise, and compression artifacts, are more favorable in improving SCMI accuracy. Code will be made available upon request.
\end{abstract}


\section{Introduction}

In recent times, devices for capturing and conveniently sharing photographs; such as smartphones, tablets, and network connected digital cameras; have become widely available and increasingly affordable, resulting in a proliferation in images. The widespread use of digital images has also underscored the need for image forensics, particularly in applications where images serve as evidence in criminal investigations and judicial proceedings. Source camera model identification (SCMI) has emerged as an important subfield of digital image forensics, where the objective is to determine the camera model that was used to capture a given image \cite{nwokeji2024source, yang2020survey}.  

Image metadata, available in EXIF headers \cite{stamm:13:access} of typical file formats such as JPEG, contains camera model information. However, such metadata is vulnerable to easy removal and manipulation, particularly in illicit use scenarios. It is crucial to develop SCMI methods that identify the camera model from the image by relying on fingerprints/traces indicative of the camera model that remain within images. SCMI methods aim to detect camera model-specific artifacts introduced during the image acquisition process and subsequent processing stages within the complex optics and image processing pipelines of camera devices. Conventional methods for SCMI have utilized features related to color filter array (CFA) demosaicing, interpolation traces, Auto-White Balance, JPEG compression, etc. Recent deep learning-based CMI methods are data-driven and rely on labeled training images to train neural networks for extracting camera model-specific features \cite{bondi2016first}, \cite{bennabhaktula2022camera}, \cite{rafi2021remnet}, \cite{rana2024dual}, \cite{ferreira2018inception}, \cite{kharrazi2004blind}. While these methods can achieve high accuracy, consistent performance across different SCMI datasets remains an issue. This work addresses this challenge by proposing SPAIR-Swin - an SCMI method based on the modified SPatial Attention with an Inverted Residual (SPAIR) block and the Swin Transformer. 

Unlike deep  learning based vision tasks, resizing images is not suitable for forensic tasks due to loss of critical information. Therefore, SCMI methods employ a patch-based strategy to extract multiple patches from an image and preserve the noise characteristics in the patch region. In this work, we propose a patch extraction strategy based on entropy that focuses on more informative patches. Entropy quantifies uncertainty and randomness, encapsulating the unique sensor noise and processing characteristics of different cameras. By prioritizing patches with higher entropy, we can extract highly discriminative camera model specific features. Furthermore, the SPAIR block enhances camera model-specific artifacts, aiding the Swin Transformer in camera model identification. Due to its sliding window attention mechanism, the Swin Transformer can focus on global information and local features of an image patch and utilizes these enhanced artifacts to perform the final classification. Extensive experiments on different SCMI datasets demonstrate that SPAIR-Swin model with our proposed entropy based patch selection strategy provides superior performance compared with prior state of the art methods. 

The main contributions of this work are as follows:
\begin{itemize}
    \item We propose SPAIR-Swin, an architecture integrating SPAIR block and Swin Transformer~\cite{liu2021swin} for robust and enhanced source camera model identification.
    \item We propose a patch extraction strategy using entropy to prioritize informative regions and improve the identification of the source camera model.
    \item We propose the SPAIR block, a unique combination of a modified spatial attention and inverted residual blocks~\cite{sandler2018mobilenetv2}, which significantly improves feature learning and identification.
    \item We conducted a comprehensive comparative study of seven different SCMI methods across four SCMI datasets.
\end{itemize}

The remainder of the paper is organized as follows. An overview of related work is present in Section \ref{sec2}. In Section \ref{sec3}, we introduce the proposed patch extraction strategy and SPAIR-Swin model. Section \ref{sec4} describes the  evaluations datasets used for the experiments. Section \ref{sec5} provides a comprehensive and extensive analysis of the performance of the proposed method and a comparison with state-of-the-art methods. Section \ref{sec6} concludes the paper.

\section{Related Works}\label{sec2}

A number of methods have been proposed for SCMI, which are surveyed in \cite{yang2020survey}. Traditional methods extract hand-crafted features from the images and perform further inference on these features. Kharrazi \textit{et al.} \cite{kharrazi2004blind} transformed each image into a numerical feature vector via extraction from the wavelet domain or computation in the spatial domain to identify cameras using supervised learning. Lukas \textit{et al.} \cite{lukas2006digital} studied sensor noise patterns in images to distinguish unique camera characteristics. Later research investigated various features, such as sensor noise statistics \cite{san2006source}, dust particles on the sensor \cite{dirik2008digital}, and combined feature sets \cite{tuama2016camera}. Many of these hand-crafted feature-based methods utilize trained support vector machines (SVMs) classifiers for SCMI based on the features. Bondi \textit{et al.} \cite{bondi2016first} developed a deep learning method using convolutional neural networks (CNN) for camera model identification (CMI). Their method consists of four convolutional layers to extract features, followed by two fully connected layers for refinement. These extracted features are then fed into linear binary support vector machines (SVMs) for final camera model classification. The final multi-class SCMI classification is realized using pairwise, one-versus-one, training of two-class SVMs. Yao \textit{et al.} \cite{yao2018robust} utilized a more intricate deep learning model comprising 11 convolutional layers to extract complex characteristics specific to individual cameras. 

Current state-of-the-art deep learning based SCMI methods employs CNN-based feature extractors followed by classification using fully connected layers. Chen \textit{et al.} \cite{chen2017camera} employed a ResNet34 model for the SCMI. Ferreira \textit{et al.} \cite{ferreira2018inception} improved CMI by combining Inception architecture and three CNNs for feature extraction. Rafi \textit{et al.} \cite{rafi2021remnet} introduced RemNet, a CNN model designed for CMI on unseen and heavily post-processed images. It includes data-adaptive pre-processing remnant blocks that adjust to image manipulations. Liu \textit{et al.} \cite{liu2021efficient} investigated the use of the Res2Net module \cite{gao2019res2net} for pre-processing, paired with a VGG16-based \cite{simonyan2014very} CMI classifier. Bennabhaktula \textit{et al.} \cite{bennabhaktula2022camera} devised a data classification system using a seven-layer Convolutional Neural Network (CNN) model. They also used a limited convolutional layer to pre-process the data. Sychandaran \textit{et al.} \cite{sychandran2024sccrnet} employed a residual network consisting of different residual networks for the SCMI. Huan \textit{et al.} \cite{huan2024camera} developed a novel patch-level compact deep network for efficient camera model identification. This network incorporates a patch extraction scheme based on a Uniform Local Binary Pattern (ULBP) to obtain complete texture information. 

Each deep learning-based SCMI method employed a different patch selection strategy designed by considering the proposed model. Bondi \textit{et al.} \cite{bondi2016first} proposed patching strategies based on mean and standard deviation, extracting 32 patches per image with a size of $64\times64$. Yao \textit{et al.}. \cite{yao2018robust} retrieved 256 patches per image of size $64\times64$ from the central $75\%$ of the image. Liu \textit{et al.} \cite{liu2021efficient} proposed a strategy to retrieve 128 patches per image of size $64\times64$, where 64 patches are calculated based on the quality measures described by Bondi \textit{et al.} \cite{bondi2016first}, and the remaining 64 are obtained using K-means clustering. Rafi \textit{et al.} \cite{rafi2021remnet} retrieved 20 patches of size $256\times256$ based on \cite{bondi2016first} and \cite{yang2019source}. Sychandran \textit{et al.} \cite{sychandran2024sccrnet} used a patch size of $32\times32$. Bennabhaktula \textit{et al.} \cite{bennabhaktula2022camera} proposed a method to extract homogeneous patches based on the standard deviation and extracted 400 overlapping patches per image with a size of $128 \times 128$. Huan \textit{et al.} \cite{huan2024camera} proposed a patching strategy using a uniform local binary pattern (ULBP) and retrieved 150 patches of size $64\times64$.

\section{Methodology}\label{sec3}

\begin{figure*}[!ht]
    \centering
    \includegraphics[width=0.9\textwidth]{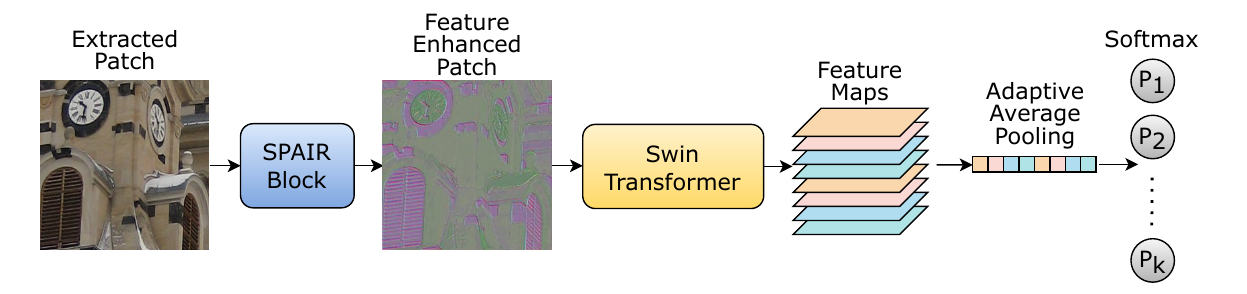}
    \caption{\text{Proposed Pipeline for SPAIR-Swin Architecture.} The process begins by extracting patches from the input image. These patches are then processed by the SPAIR block, which generates Feature-Enhanced Patches that capture prominent visual cues. The enhanced patches are subsequently fed into the Swin Transformer for the final classification.}
    \label{fig:proposedarch}
\end{figure*}

The proposed pipeline incorporating the entropy-based patching strategy, the SPAIR block, and the SPAIR-Swin architecture for source camera model identification is illustrated in Fig.~\ref{fig:proposedarch}. Individual modules in the pipeline are described in the following subsections.

\subsection{Entropy based Patch Extraction}

In deep learning applications, before processing, images are typically resized to a fixed size that can be handled efficiently using the parallelism available in hardware. Although resizing preserves semantic content for vision-related tasks, it is sub-optimal for forensics, including SCMI, because it can suppress or eliminate subtle features indicative of the camera model that are unrelated to the image semantic content. Instead of scaling, SCMI methodologies therefore use fixed size patches cropped from the input image for use in deep learning pipelines. The selection of image regions for the extraction of patches plays a critical role in determining the performance of SCMI algorithms. A good selection strategy can better isolate regions carrying camera model information and enhance the generalization performance. 

\begin{figure}[!ht]
    \centering
    \includegraphics[width=\columnwidth]{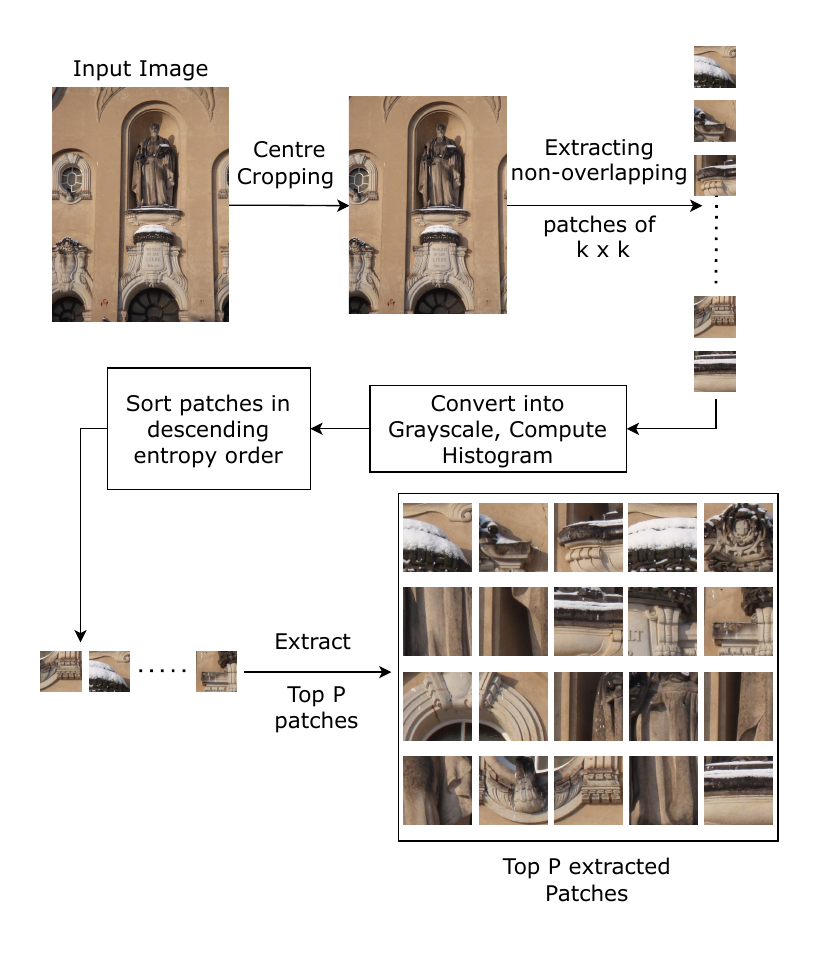}
    \caption{\text{Entropy-based patch extraction.} The input images are first center-cropped, then processed to get non-overlapping patches of size $k \times k$. For each patch, entropy is computed based on its grayscale pixel intensity distribution. The patches are ranked in descending order of entropy and the top $\mathbf{P}$ patches are selected for the purpose of SCMI.}
    \label{fig:patching}
\end{figure}

We propose a patch extraction strategy based on entropy, which is illustrated in Fig.~\ref{fig:patching}. By partitioning the images into non-overlapping patches, the model can focus more on the regions with higher information content that contain sensor noise patterns like PRNU noise, chromatic aberrations, and demosaicing artifacts. We argue these high-entropy regions are likely to hold various textures and patterns well-suited for source camera identification.

Areas with higher entropy correlate to regions containing variations in pixel intensity and often such areas contain sensor-specific noise patterns relevant for identifying cameras. Entropy based selection preferably retains patches with high frequency noise that are critical for SCMI, in contrast with traditional discrete cosine transforms (DCT) based JPEG compression methods that tend to discard high-frequency noise. By concentrating on non-homogeneous patches that capture the most discriminative features, our method facilitates the extraction of robust camera fingerprints.

Specifically, our patch selection is based on the (first-order) entropy is computed for an $8$-bit grayscale image patch $X$ as 
\begin{equation}\label{eq:1}
H(X) = -\sum_{x=0}^{255} p(x) \cdot \log(p(x))
\end{equation} 
where,  $p(x)$ denotes the fraction of pixels in the patch $X$ having a grayscale intensity $x$. 
First, the source image is center-cropped to maintain the overall composition of the image. The purpose of center cropping is to ensure the image is evenly divisible by the patch size, preserving the central content while removing peripheral details from the boundary areas.Then, non-overlapping patches are extracted by using a $k \times k$ sliding window with a stride of $k$ pixels along horizontal and vertical directions. A patch size of $k=256$ is used as a balanced choice to capture global and local features. This size maintains crucial information and is suitable for the SWIN Transformer architecture. Following the patch extraction, we compute the entropy of each patch using Equation~\eqref{eq:1}.  Patches are then sorted in descending order of entropy and the $P$ patches with the highest entropy are chosen. Empirically, the parameter $P$ is determined to optimize the trade-off between information richness and computational requirements. In Section~\ref{sec5}, we analyze the effect of different values of $P$ on model performance and determine the optimal configuration that maximizes the discriminative capability of our model without adding redundancy.


\subsection{Proposed SPAIR Block for Enhanced Feature Extraction}

To extract features useful for SCMI from individual image patches, we proposed a SPAIR block that combines an Inverted Residual Block~\cite{sandler2018mobilenetv2} with a modified spatial attention block~\cite{woo2018cbam}. The SPAIR block makes use of spatial attention to focus on distinctive features of interest by amplifying camera-specific features whereas the Inverted Residual Block utilizes its expansion layer and depthwise convolutions to retain information and to focus on unique channel-wise patterns.

\begin{figure*}[!ht]
    \centering
    \includegraphics[width=0.9\textwidth]{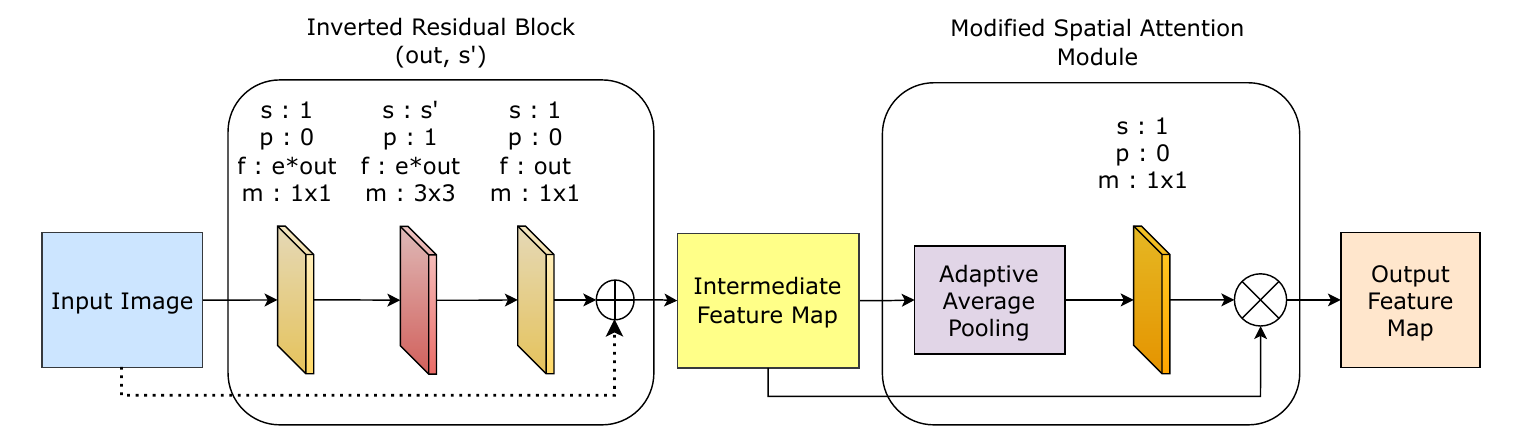}
    \caption{\text{The schematic diagram of the SPAIR module.} The proposed feature extractor incorporates an Inverted Residual Block with an expansion factor of $4$ and a modified spatial attention module. The Inverted Residual Block makes use of depthwise convolution coupled with elementwise additive skip connection, while the modified spatial attention module uses average pooling with an adaptive mechanism to improve feature representation. ReLU and batch normalization are added after convolutional and depthwise convolutional layers except for the last layer that employs a convolutional layer with a sigmoid function. \textbf{s}: stride, \textbf{p}: padding, \textbf{f}: number of filters, \textbf{m}: kernel size or mask size, \textbf{e}: expansion factor, $\mathbf{\otimes}$: elementwise multiplication, $\mathbf{\oplus}$: elementwise addition.}
    \label{fig:FE}
\end{figure*}

The Inverted Residual Block~\cite{sandler2018mobilenetv2} is commonly used in many deep convolutional neural networks (CNNs). Its lightweight nature helps to create feature maps efficiently as it utilizes point-wise convolutions for dimensionality reduction and expansion. This is followed by a depth-wise separable convolution that significantly reduces the computational cost compared to the standard convolution. This combination promotes efficient feature learning while maintaining spatial information. The initial input is added to the output of the second convolution layer before passing it to modified spatial attention module. This residual connection helps with the gradient flow during training.

Unlike traditional modified spatial attention module~\cite{woo2018cbam}, which uses both Max Pooling and Average Pooling, we simplify the process by focusing only on Average Pooling to reduce the redundancy while still capturing global features. The softmax layer utilized at the end ensures that attention features are globally distributed across the entire spatial dimension, enabling our model to focus on the most discriminative features. This targeted attention mechanism helps to suppress irrelevant background content and focus on global noise patterns that differentiate the camera models. By selectively improving informative regions, SPAIR aims to extract more discriminative features that effectively distinguish images captured by different cameras.


\subsection{Swin Transformer}

\begin{figure*}[!ht]
    \centering
    \includegraphics[width=\textwidth]{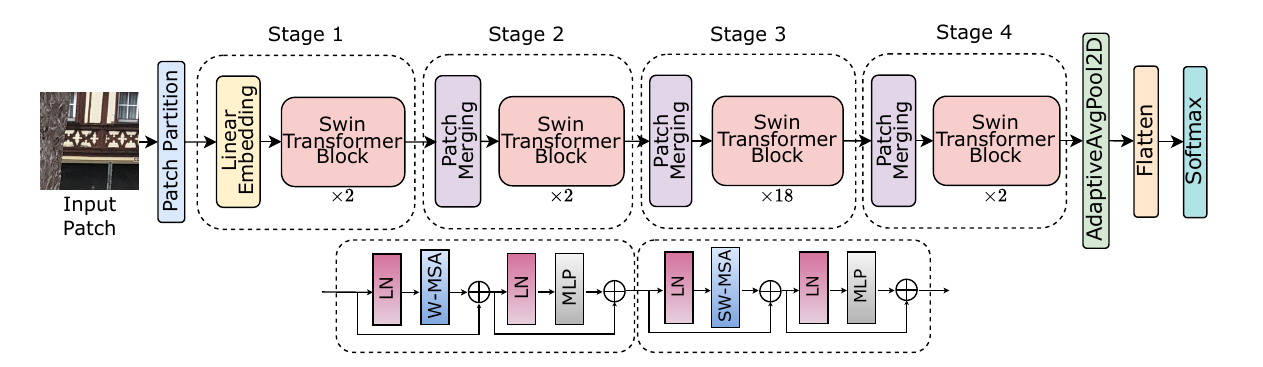}
    \vspace{2mm}
    \caption{Swin Transformer Architecture. The components of the Swin Transformer, including the LayerNorm (LN) layer, MultiLayer Perceptron (MLP), and Multi-Head Self-Attention modules (W-MSA and SW-MSA), with regular and shifted windowing configurations, respectively. }

    \vspace{3mm}
    \label{fig:swint}
\end{figure*}

The output of the SPAIR block is processed by a Swin Transformer~\cite{liu2021swin}. The Swin Transformer is a hierarchical transformer model designed specifically for computer vision/image processing that is more efficient than the initial proposals for vision transformers. The embedded patches are converted to tokens that form the base of the input to the transformer. Positional encoding is added to the tokens to supply spatial information for the model to be aware of the position of patches relative to each other. Another unconventional characteristic of the Swin Transformer is the low-cost manner in which it accommodates multi-scale features. Tokens at lower levels represent fine-grained details, whereas tokens at higher levels capture more abstract and global features. The final feature maps are obtained by combining all tokens which gives an overall view of the input features. These features obtained are then entered into the classification head.

\section{Experimental Setup: Datasets, Metrics, and Baseline} \label{sec4}

\subsection{Datasets}\label{sec:dataset}
To assess the robustness and generalizability of the proposed SCMI method, we perform experiments on four large publicly available datasets: Dresden \cite{gloe2010dresden}, Vision \cite{shullani2017vision}, Forchheim \cite{hadwiger2021forchheim}, and Socrates \cite{galdi2019socrates}. The selection of smartphone image datasets is significant as most images these days are acquired from smartphone cameras. Table \ref{tab:dataset} shows a summary of the datasets used. Before discussing the experiments, we outline the pertinent details of each of these datasets:

\begin{itemize}
      \item The Dresden dataset \cite{gloe2010dresden} is the largest with respect to the number of images used in our experiments. It comprises $14999$ images captured by $18$ camera models. As indicated in \cite{kirchner2015forensic}, Nikon\_D70 and Nikon\_D70s have been consolidated into a single camera model. 
    \item The Vision dataset \cite{shullani2017vision} includes images captured with various smartphone cameras. It comprises $11732$ original images distributed over $35$ camera devices and contains $5$ identical camera models. We combine these classes for our experiments.
    \item The Forchheim dataset \cite{hadwiger2021forchheim} includes $143$ scenes taken with $9$ different brands of smartphones. It has $3851$ images taken under various conditions, which include changes in lighting parameters, ambient light, and imaging parameters. Two out of the $27$ camera devices are identical, indicating that two models have more than one device, which are merged for Source Camera Model Identification tests. Consequently, there are $25$ different camera devices. This study focuses on native camera images that are labeled \textit{original} (orig.) in \cite{hadwiger2021forchheim}.
    \item The Socrates \cite{galdi2019socrates} is the largest dataset in terms of number of camera devices and models, containing $9721$ images taken with $65$ different camera models. This dataset offers a rich variety of scenarios and authenticity, with images captured by individual smartphone users.
\end{itemize}

\begin{table}[!ht]  
    \caption{Overview of datasets utilized in this study.}
    \vspace{3mm}
    \centering    
    \begin{tabularx}{\columnwidth}{lYY}
        \toprule
        \textbf{Dataset} & \textbf{Number of Images} & \textbf{Number of Camera Models} \\ \midrule
        Dresden~\cite{gloe2010dresden} & 14999 & 18 \\
        Forchheim~\cite{hadwiger2021forchheim} & 3851 & 25 \\
        Vision~\cite{shullani2017vision} & 11732 & 30 \\ 
        Socrates~\cite{galdi2019socrates} & 9721 & 65 \\ \bottomrule
    \end{tabularx}
    \label{tab:dataset}
\end{table}


\subsection{Setup}

All the experiments are performed on a $3.38$ GHz AMD EPYC $7742$, $64$ core processor with $64$ GB RAM and an Nvidia A$100$ GPU with $40$ GB memory. Images in the dataset are split in an $80:20$ ratio. Splitting is performed before patching to ensure that the derived patches are present in either the training or the test set. The PyTorch \cite{paszke2019pytorch} framework was used for implementation. A pre-trained Small Swin Transformer \cite{liu2021swin} is used that has a batch size of $32$. AdamW \cite{loshchilov2017decoupled} optimizer is used that has an initial learning rate of $0.0001$. All models were trained for $100$ epochs, at which point, they exhibited convergence in training loss, except for our proposed model which showed convergence around $50$ epochs.

\subsection{Evaluation Metrics}

To evaluate the proposed method, we measured Image Level Accuracy (ILA) and Patch Level Accuracy (PLA). The evaluation was carried out with a total of $N$ images collected by $C$ camera devices. For the $i^{th}$ image, $X_i$, $Y_i$, and $\hat{Y}_i$ denote the specific image, the actual camera model, and the predicted camera model, respectively. Furthermore, the $\mathcal{P}_i$ number of patches has been extracted from the $i^{th}$ image. For the $j^{th}$ patch of the $i^{th}$ image, $y_{ij}$ and $\hat{y}_{ij}$ represent the actual and predicted camera models, respectively, for that patch image. PLA and ILA are computed as
\begin{equation*}
    PLA = \frac{\sum_{i=1}^{N}\sum_{j=1}^{\mathcal{P}_i}\mathds{1}(\hat{y}_{ij}=y_{ij})}{\sum_{i=1}^{N}\mathcal{P}_i}
\end{equation*}

\begin{equation*}
    ILA=  \frac{\sum_{i=1}^{N}\mathds{1}(\hat{Y}_{i}=Y_{i})}{N}
\end{equation*}
where $\mathds{1}(.)$ is the indicator function and $\hat{Y}_i$ corresponds to the camera model with the maximum number of votes according to the patch-level predictions of the image and its mathematical formulation is $\hat{Y}_i = \argmax{k}(\sum_{j=1}^{\mathcal{P}_i}\mathds{1}(\hat{y}_{ij}=k))$.

For comprehensive analysis across the datasets, we also used F1-score where unweighted mean was used to compute a single value for ease of understanding and visualization.

\subsection{Baselines}\label{sec:baselines}

We compare the performance of the proposed method against several competing methods for Camera Model Identification (CMI), specifically
those proposed by Chen \textit{et al.} \cite{chen2017camera}, Liu \textit{et al.} \cite{liu2021efficient}, Rafi \textit{et al.} \cite{rafi2021remnet}, Bennabhaktula \textit{et al.} \cite{bennabhaktula2022camera}, Sychandran \textit{et al.} \cite{sychandran2024sccrnet}, and Huan \textit{et al.} \cite{huan2024camera}. Chen \textit{et al.} \cite{chen2017camera} employed a CNN-based method without pre-processing strategies while all other SCMI methods utilized varied pre-processing strategies to enhance image features. This variety of methodological frameworks enhances comparative analysis by highlighting the strengths and limitations of each method in the context of source camera model identification.

\section{Results and Analysis}\label{sec5}

The performance of the proposed SPAIR-Swin method is examined in detail.  For the comparative analysis, extensive experiments were conducted to compare the proposed approach with the baselines outlined in Section \ref{sec:baselines}, utilizing four distinct datasets detailed in Section \ref{sec:dataset}.

\begin{table*}[!ht]
    \caption{Comparison of image-level accuracy (ILA) and patch-level accuracy (PLA) for four camera model identification methods on four datasets. SPAIR-Swin outperforms previous methods with higher accuracy in both ILA and PLA on all datasets.}
    \vspace{3mm}
    \centering
    \begin{tabularx}{\textwidth}{lYYYYYYYY}
        \toprule
        \multirow{2.1}{*}{\begin{tabular}[c]{@{}l@{}}\textbf{Dataset} $\rightarrow$\\ \textbf{Methods} $\downarrow$\end{tabular}}  & \multicolumn{2}{c}{\textbf{Dresden \cite{gloe2010dresden}}} & \multicolumn{2}{c}{\textbf{Vision \cite{shullani2017vision}}} & \multicolumn{2}{c}{\textbf{Forchheim \cite{hadwiger2021forchheim}}} & \multicolumn{2}{c}{\textbf{Socrates \cite{galdi2019socrates}}} \\
        & \textbf{ILA} & \textbf{PLA} & \textbf{ILA} & \textbf{PLA} & \textbf{ILA} & \textbf{PLA} & \textbf{ILA} & \textbf{PLA} \\ \midrule
        {Chen \textit{et al.}} \cite{chen2017camera}   & 94.33 & -   & 87.48 & -   & 48.91 & -   & 77.31 & - \\
        {Liu \textit{et al.}} \cite{liu2021efficient}  & 97.30 & 92.63 & 95.33 & 86.52 & 97.19 & 85.37 & 96.25 & 88.59 \\ 
       {Rafi \textit{et al.}} \cite{rafi2021remnet}  & 99.42 & 96.16 & 98.73 & 92.38 & 99.23 & 89.55 & 95.89 & 84.36 \\ 
       {Bennabhaktula \textit{et al.}}  \cite{bennabhaktula2022camera}   & 98.76 & 96.96 & 97.71 & 93.53 & 95.02 & 90.30 & 93.06 & 87.50 \\ 
        {Sychandran \textit{et al.}} \cite{sychandran2024sccrnet}  & 94.29 & 82.00 & 79.47 & 65.00 & 63.35 & 47.03 & 43.94 & 38.72 \\ 
        {Huan \textit{et al.}} \cite{huan2024camera} & 97.65   & 95.10 & 95.38 & 90.47 & 93.87 & 83.74 & 93.43 & 86.35 \\ 
       {Proposed} & \textbf{99.87} & \textbf{99.45} & \textbf{99.32} & \textbf{98.39} & \textbf{100} & \textbf{99.45} & \textbf{98.61} & \textbf{97.46} \\ \bottomrule
    \end{tabularx}
    \label{table:performance}
\end{table*}

\begin{figure}[!ht]
    \centering
    \includegraphics[width=\columnwidth]{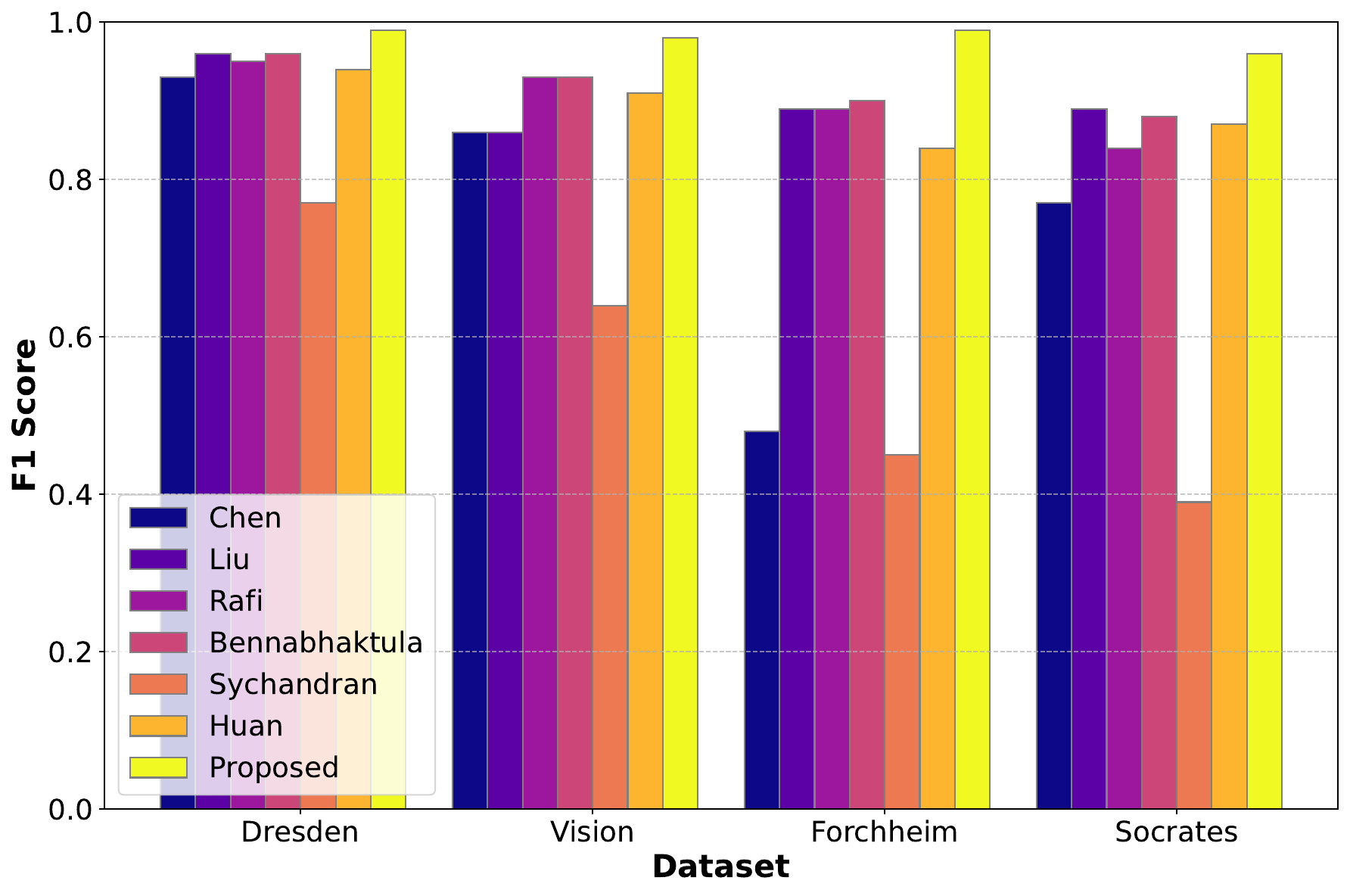}
    \caption{Comparison of F1 Scores achieved by various methods on the four datasets. Across all datasets, the proposed method has outperformed others and achieved almost perfect F1 scores in most cases, indicating its generalizability across different datasets. Previous state-of-the-art methods displayed some inconsistencies in their performance, where few of them experiencing drastic reductions in their F1 scores.}
    \label{fig:performance}
\end{figure}

 Table \ref{table:performance} presents the performance, in terms of ILA and PLA, of various Camera Model Identification (CMI) methods, including our proposed approach. The best scores are highlighted in bold and it is noted that the proposed approach outperforms all competing methods across all four datasets. Our proposed method achieves an average absolute improvement over baseline methods in terms of ILA by $2.91\%$, $6.97\%$, $17.07\%$, and $15.30\%$, and in terms of PLA by $6.88\%$, $12.81\%$, $20.25\%$, and $20.36\%$ on the Dresden~\cite{gloe2010dresden}, Vision~\cite{shullani2017vision}, Forchheim~\cite{hadwiger2021forchheim}, and Socrates~\cite{galdi2019socrates} datasets, respectively. Furthermore, Figure \ref{fig:performance} presents a comparative analysis of patch-level F1-scores across four datasets, demonstrating that the proposed method improves classification performance by achieving a higher F1-score and reducing incorrect predictions. These results highlight the effectiveness of the proposed approach. 

\subsection{Validation of Patch Selection Strategy}
This section provides a comprehensive analysis in order to demonstrate the efficacy of the proposed patch selection strategy.
The key motivation for selecting non-homogeneous patches, i.e., high-entropy patches, is their ability to capture camera-specific features that are unevenly distributed across the image. These features include sensor noise, demosaicing patterns, and compression signatures, which are prominently found in parts with complex textures, edges, and high-frequency component areas. \par

\begin{table*}[!ht]
\caption{Comparison of ILA and PLA scores of various state-of-the-art camera model identification methods, considering both - their native  patch extraction strategy and also the the patch extraction strategy proposed in this paper. When applying the proposed patch extraction strategy, significant increases in ILA and PLA have been observed for all methods for almost all the four datasets considered. The enhanced results imply that our method delivers better discriminative and informative patches, thus significantly improving the effectiveness of existing schemes.}
\vspace{3mm}
\centering
\resizebox{\textwidth}{!}{
\begin{tabular}{lcccccccc}
\toprule
\multirow{2}{*}{\begin{tabular}[c]{@{}l@{}}\textbf{Dataset} $\rightarrow$\\ \textbf{Methods} $\downarrow$\end{tabular}}  & \multicolumn{2}{c}{\textbf{Dresden \cite{gloe2010dresden}}} & \multicolumn{2}{c}{\textbf{Vision \cite{shullani2017vision}}} & \multicolumn{2}{c}{\textbf{Forchheim \cite{hadwiger2021forchheim}}} & \multicolumn{2}{c}{\textbf{Socrates \cite{galdi2019socrates}}} \\ 
\textbf{Methods} & {Original} & {Proposed} & {Original} & {Proposed} & {Original} & {Proposed} & {Original} & {Proposed} \\ \midrule
 \multicolumn{9}{c}{{Patch Level Accuracy (PLA)}} \\ \midrule
{Liu \textit{et al.}} \cite{liu2021efficient} & 92.63 & \textbf{96.81} & 86.52 & \textbf{93.58} & 85.37 & \textbf{94.09} & 88.59 & \textbf{91.78} \\ 
Rafi \textit{et al.}~\cite{rafi2021remnet} & 96.16 & \textbf{97.12} & 92.38 & \textbf{92.42} & 89.55 & \textbf{91.90} & 84.36 & \textbf{87.95} \\ 
Bennabhaktula \textit{et al.}~\cite{bennabhaktula2022camera} & 96.96 & \textbf{97.54} & 93.53 & \textbf{94.57} & 90.30 & \textbf{91.61} & 87.50 & \textbf{91.78} \\ 
Sychandran \textit{et al.}~\cite{sychandran2024sccrnet} & 82.00 & \textbf{85.49} & 65.00 & \textbf{65.77} & 47.03 & \textbf{62.81} & 38.72 & \textbf{38.96} \\ 
Huan \textit{et al.}~\cite{huan2024camera} & 95.10 & \textbf{99.18} & 90.47 & \textbf{97.47} & 83.74 & \textbf{98.01} & 86.35 & \textbf{94.16} \\  \midrule
 \multicolumn{9}{c}{{Image Level Accuracy (ILA)}} \\ \midrule
{Liu \textit{et al.}} \cite{liu2021efficient} & 97.30 & \textbf{98.10} & 95.33 & \textbf{97.45} & 97.19 & \textbf{98.85} & 96.25 & \textbf{96.56} \\ 
Rafi \textit{et al.}~\cite{rafi2021remnet} & 99.42 & \textbf{99.78} & \textbf{98.73} & 98.39 & 99.23 & \textbf{99.49} & 95.89 & \textbf{96.36} \\  
Bennabhaktula \textit{et al.}~\cite{bennabhaktula2022camera} & 98.76 & \textbf{99.56} & 97.71 & \textbf{98.26} & 95.02 & \textbf{98.34} & 93.06 & \textbf{96.36} \\  
Sychandran \textit{et al.}~\cite{sychandran2024sccrnet} & 94.29 & \textbf{96.81} & 79.47 & \textbf{82.61} & 63.35 & \textbf{83.91} & 43.94 & \textbf{44.82} \\  
Huan \textit{et al.}~\cite{huan2024camera} & 97.65 & \textbf{99.96} & 95.38 & \textbf{99.27} & 93.87 & \textbf{100} & 93.43 & \textbf{98.05} \\  \bottomrule
\end{tabular}}
\label{tab:comparison}
\end{table*}

Table~\ref{tab:comparison} presents the comparative results of various baseline methods experimented across four different datasets with their respective native patch selection strategy and the proposed patch selection strategy while keeping all the experimental settings consistent with the original studies, which include the model and training pipelines, their respective hyperparameters, the number of patches, and patch sizes. The results demonstrate the effectiveness of the proposed entropy-based patch extraction in enhancing camera model identification performance across various state-of-the-art methods. It can be observed, baseline methods with our proposed patch selection strategy achieve an average relative PLA improvement of $2.93\%$, $3.65\%$, $12.98\%$, and $4.48\%$ , and an average relative ILA improvement of $1.41\%$, $2.09\%$, $8.89\%$, and $2.26\%$ over their native patch selection strategy on the Dresden~\cite{gloe2010dresden}, Vision~\cite{shullani2017vision}, Forchheim~\cite{hadwiger2021forchheim}, and Socrates~\cite{galdi2019socrates} datasets, respectively. Notably, all studies show improvements in both PLA and ILA across all the datasets, except for a marginal ILA loss of $0.34\%$ in Liu et al.\cite{liu2021efficient}. The relative gains are massive especially in calibration using difficult datasets like Forchheim which contains a high amount of similar images, and Socrates which has fewer samples per camera model and the highest classes for identification. For example, the method developed by Huan \textit{et al}., experiences a drastic change in PLA from $83.74\%$ to $98.01\%$ for the Forchheim~\cite{hadwiger2021forchheim} data set and from $86.35\%$ to $94.16\%$ of the Socrates~\cite{galdi2019socrates} dataset. These improvements are statistically significant across all the methodologies and datasets considered, demonstrating the consistency and expansiveness of the proposed patch extraction strategy.

\begin{table}[!ht]
\caption{PLA and ILA comparison between homogeneous and non-homogeneous patches. The heterogeneous patches consistently surpass their homogeneous counterparts in terms of PLA over all data sets. Specifically, the Socrates dataset has shown the highest improvement where it recorded an increase in accuracy by 1.53\% for ILA.}
\vspace{3mm}
\centering
\label{table:patchcomp}
\begin{tabular}{lcc}
\toprule
\textbf{Dataset} & \textbf{Homogeneous} & \textbf{Non Homogeneous} \\ \midrule
\multicolumn{3}{c}{Patch Level Accuracy (PLA)} \\ \midrule
    Dresden~\cite{gloe2010dresden} & 99.25 & \textbf{99.45} \\ 
    Vision~\cite{shullani2017vision} & 97.90 & \textbf{98.39} \\ 
    Forchheim~\cite{hadwiger2021forchheim} & 98.81 & \textbf{99.45} \\ 
    Socrates~\cite{galdi2019socrates} & 96.25 & \textbf{97.46} \\  \midrule
\multicolumn{3}{c}{Image Level Accuracy (ILA)} \\ \midrule
    Dresden~\cite{gloe2010dresden} & \textbf{99.91} & 99.87 \\ 
    Vision~\cite{shullani2017vision} & 98.81 & \textbf{99.32} \\ 
    Forchheim~\cite{hadwiger2021forchheim} & 99.87 & \textbf{100} \\ 
    Socrates~\cite{galdi2019socrates} & 97.12 & \textbf{98.61} \\ \bottomrule
\end{tabular}
\end{table}

Table~\ref{table:patchcomp} shows the performance of the proposed CMI method by considering both scenarios: (1) using homogeneous patches and (2) using  non-homogeneous patches.  The higher patch-level accuracy is achieved by using the non-homogeneous patches, as their high-entropy regions contain rich textures, edges, and structural details, effectively captured by the shifted window multi-head self-attention module of Swin Transformers. The modified spatial attention mechanism assigns weights to emphasize distinctive regions, aiding camera-specific feature extraction. In contrast, homogeneous patches, being uniform, offer fewer distinguishing features, making artifact isolation for SCMI more challenging. Moreover, the improvement in image-level precision, except for a marginal decline on Dresden~\cite{gloe2010dresden}, highlights the advantage of high-entropy patches.

\begin{table}[!ht]
    \caption{PLA and ILA comparison study for different patch sizes (15, 20, 30, and 60). The highest accuracy is obtained when the patch size is set to 20. This optimal patch size is thus selected for inclusion in the pipeline to balance performance and computational efficiency.}
    \vspace{3mm}
    \centering
    \begin{tabular}{lcccc}
    \toprule
    \multirow{2}{*}{\begin{tabular}[c]{@{}l@{}}\textbf{\# Patches}$\rightarrow$ \\ \textbf{Dataset} $\downarrow$\end{tabular}} & \multirow{2}{*}{\textbf{15}} & \multirow{2}{*}{\textbf{20}} & \multirow{2}{*}{\textbf{30}} & \multirow{2}{*}{\textbf{60}} \\ 
    & & & & \\
    \midrule \multicolumn{5}{c}{{Patch Level Accuracy (PLA)}} \\ \midrule
    {Dresden~\cite{gloe2010dresden}}  & 99.18 & \textbf{99.45} & 99.23 & 99.19 \\ 
    {Vision~\cite{shullani2017vision}} & 97.74 & \textbf{98.39} & 98.22 & 98.35 \\ 
    {Forchheim~\cite{hadwiger2021forchheim}} & 98.67 & \textbf{99.45} & 99.20 & 99.18 \\ 
    {Socrates~\cite{galdi2019socrates}} & 96.57 & \textbf{97.46} & 96.96 & 96.97 \\ 
     \midrule \multicolumn{5}{c}{{Image Level Accuracy (ILA)}} \\ \midrule
    {Dresden~\cite{gloe2010dresden}}  & 99.82 & {99.87} & \textbf{99.91} & 99.68 \\ 
    {Vision~\cite{shullani2017vision}} & 99.19 & \textbf{99.32} & 98.89 & 99.19 \\ 
    {Forchheim~\cite{hadwiger2021forchheim}} & 99.74 & \textbf{100.00} & 99.87 & 99.61 \\ 
    {Socrates~\cite{galdi2019socrates}} & 97.63 & \textbf{98.61} & 98.35 & 98.45 \\ \bottomrule
    \end{tabular}
    \label{tab:patchresults}
\end{table}

An ablation analysis was conducted to determine the optimal number of patches for maximizing the proposed model's performance, while addressing underfitting and overfitting. These patches must also be non-redundant about the information they convey as well as the patches should be high in quality. Table \ref{tab:patchresults} shows that increasing the number of patches from 15 to 20 consistently improved both patch-level accuracy (PLA) and image-level accuracy (ILA) across four datasets. Overall, the best results are achieved when the number of high-entropy patches selected per image is 20. Increasing the number of such patches to 30 or 60 results in  performance drop in most of the cases. 

\section{Conclusion}\label{sec6}

This paper introduced a novel Source Camera Model Identification (SCMI) framework, utilizing SPAIR-Swin which integrates SPAIR block with the Swin Transformer. The architecture effectively captures global and local features present in the image and provide a robust SCMI method. The use of the entropy-based patch selection enables the model to concentrate on areas of higher information content (higher entropy) and thus discriminate between discriminative camera-specific features such as sensor noise and demosaicing artifacts. Spatial Attention Inverted Residual Blocks (SPAIR) further boost the learning of features by combining it with the Swin Transformer, which contains efficient hierarchical self-attention adapted for robust classification. The experiments highlighted the superior performance of our method on four publicly available standard datasets, achieving state-of-the-art accuracy in both patch level (PLA) and image level (ILA). We demonstrated that our entropy-based patch extraction strategy worked well with most existing SCMI methods, and ablation studies signified the efficacy of the proposed method. We confirm the potential of the SPAIR-Swin model as a powerful method for SCMI, pointing the way toward future research in digital image forensics.

\bibliographystyle{IEEEbib}
\bibliography{main}

\end{document}